\documentclass[journal]{IEEEtran}
\usepackage{cite,graphicx,subfigure,threeparttable,booktabs,amsmath,amsfonts,amssymb,bm}
\usepackage{caption2}
\usepackage{verbatim}
\usepackage{balance}
\usepackage{color}
\usepackage{multirow}
\usepackage{color}
\ifCLASSINFOpdf
\else
\fi

\begin{document}

\title{Saliency Detection with Spaces of Background-based Distribution}

\author{Tong Zhao, Lin Li, Xinghao Ding, Yue Huang and Delu Zeng*
\thanks{Accepted by IEEE Signal Processing Letters in Mar. 2016. D. Zeng, T. Zhao, L. Li, Y. Huang and X. Ding are with School of Information Science and Engineering, Xiamen University, China.}
\thanks{This work was supported in part by the National Natural Science Foundation of China under Grants 61103121, 61172179, 81301278, 61571382 and 61571005, in part by the Fundamental Research Funds for the Central Universities under Grants 20720160075, 20720150169 and 20720150093, in part by the Natural Science Foundation of Guangdong Province of China under Grants 2015A030313589 and by the Research Fund for the Doctoral Program of Higher Education under Grant 20120121120043.}
\thanks{*Corresponding to: dltsang@xmu.edu.cn for Delu Zeng.}
}

\maketitle

\begin{abstract}
In this letter, an effective image saliency detection method is proposed by constructing some novel spaces to model the background and redefine the distance of the salient patches away from the background. Concretely, given the backgroundness prior, eigendecomposition is utilized to create four spaces of background-based distribution (SBD) to model the background, in which a more appropriate metric (Mahalanobis distance) is quoted to delicately measure the saliency of every image patch away from the background. After that, a coarse saliency map is obtained by integrating the four adjusted Mahalanobis distance maps, each of which is formed by the distances between all the patches and background in the corresponding SBD. To be more discriminative, the coarse saliency map is further enhanced into the posterior probability map within Bayesian perspective. Finally, the final saliency map is generated by properly refining the posterior probability map with geodesic distance. Experimental results on two usual datasets show that the proposed method is effective compared with the state-of-the-art algorithms.
\end{abstract}


\IEEEpeerreviewmaketitle

\section{Introduction}

\IEEEPARstart{A}{s} a subfield of computer vision and pattern recognition, image saliency detection is an interesting and challenging problem which touches upon the knowledge from multiple subjects like mathematics, biology, neurology, computer science and so on. Many previous works have demonstrated that saliency detection is significant in many fields including image segmentation \cite{li2011saliency}, objection detection and recognition \cite{ren2014region}, image quality assessment \cite{li2013color}, image editing and manipulating \cite{margolin2013saliency}, visual tracking \cite{zhang2010visual}, etc.

The models of saliency detection can be roughly categorized into bottom-up and top-down approaches \cite{borji2014salient},\cite{borji2015salient}. In this work, we focus on bottom-up saliency models. In recent years, color contrast cue \cite{cheng2015global},\cite{liu2014saliency} and color spatial distribution cue \cite{perazzi2012saliency} are widely used in saliency detection. It is considered that salient objects always show large contrast over their surroundings and are generally more compact in the image domain. Saliency algorithms based on these two cues usually work well when the salient object can be distinctive in terms of contrast and diversity, but they usually fail in processing images containing complicated background which involves various patterns. On the other hand, pattern distinctness is described in \cite{margolin2013makes}  by studying the inner statistics (actually corresponds to the distribution for the feature) of the patches in PCA space. This method achieves favorable results but there still exists some background patterns which are misjudged to be salient. In addition,  by assuming most of the narrow border of the image be the background region,  backgroundness prior plays an important role in calculating the saliency maps in many approaches \cite{li2013saliency},\cite{yang2013saliency},\cite{wei2012geodesic}. However, if the patches on the border are treated as background patches directly, these algorithms often perform inadequately when the potential salient objects touch the image border.

Based on the above analysis, we propose a saliency detection method considering the following aspects which also go to our contributions: 1) We combine the backgroundness prior into constructing the spaces of background-based distribution  implicitly (not directly) to investigate the background, where we will find Euclidean and $l_1$ metrics may not be suitable in exploring the distribution (inner statistics) for the patches and we should quote some other metric to describe the patch difference appropriately. 2) In the step of refinement, we also design a new up-sampling scheme based on geodesic distance.

\section{Proposed algorithm}

\begin{figure}[t]
\centering
\subfigure[]
{\includegraphics[width=0.65in]{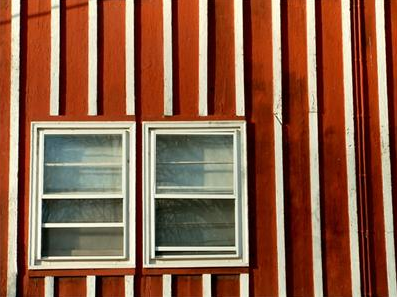}}
\subfigure[]
{\includegraphics[width=0.65in]{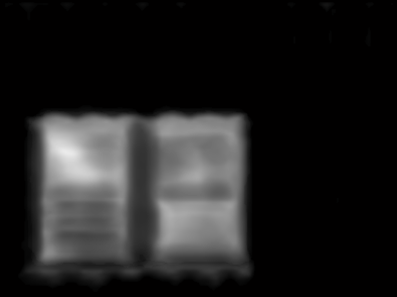}}
\subfigure[]
{\includegraphics[width=0.65in]{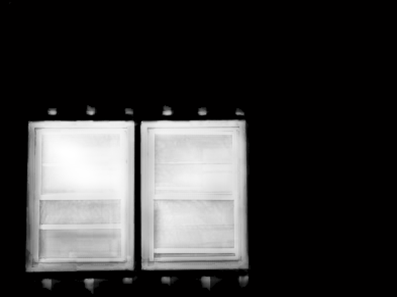}}
\subfigure[]
{\includegraphics[width=0.65in]{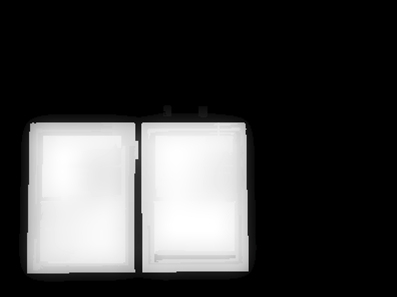}}
\subfigure[]
{\includegraphics[width=0.65in]{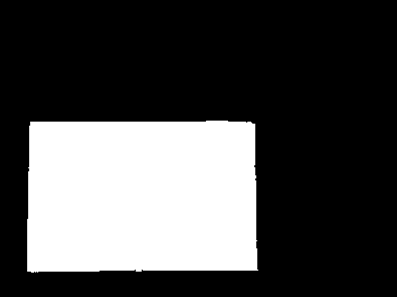}}
\vspace{-7pt}
\caption{Main framework of the proposed approach. (a) Input image. (b) Coarse saliency map based on SBD. (c) Posterior probability map based on Bayesian perspective. (d) Refinement. (e) Ground-truth.}
\label{fig.1}
\end{figure}

Here we describe the proposed approach in three main steps. Firstly, the patches from image border are used to generate a group of spaces of background-based distribution (SBD) to compute the coarse saliency map.  Secondly, in the Bayesian perspective, the coarse saliency map is enhanced. Then the final saliency map is obtained by using a novel up-sampling method based on geodesic distance. And the main framework is depicted in Fig. \ref{fig.1}.

\subsection{Spaces of background-based distribution}	
The work \cite{margolin2013makes} suggests that a pixel be salient if the patch containing this pixel has less contribution to the distribution of the whole image treated in PCA system. However, problem may occur when the area of the salient object is overwhelming in size. In this case, if all the image patches are utilized to do PCA, the patches in the salient object may have high contribution to distribution of the whole image patches, then naturally the difference between the salient patches and the whole image will not be discriminative. In order to increase this difference and make the salient object be much discriminative from the background, in this paper we aim to re-establish the background distribution in some new spaces, called spaces of background-based distribution (SBD). For this purpose, we have the following considerations:

 \textbf{1) Feature representation} For the patches, color information is used to represent their features. Particularly, two color spaces are used here, including CIE-Lab and I-RG-BY. CIE-Lab is most widely used in previous works, while I-RG-BY that is transformed from RGB color space is also proved to be effective in saliency detection algorithm \cite{frintrop2015traditional}, where $I=\frac{R+G+B}{3}$ denotes the intensity channel, $RG=R-G$ and $BY=B-\frac{R+G}{2}$ are two color channels. Thus, if we take a $7\times 7$ patch as an example, the feature of this patch will be represented by a $294\times 1$ column vector $\left (6\times 7\times 7=294\right )$. And here we denote this vectorized feature for image patch $i$ to be $f(i)$.

\textbf{2) Background patches selection} We define the pseudo-background region as the border region of the image like the work in \cite{jiang2013salient}. To improve the performance, some approaches \cite{yang2013saliency},\cite{han2014background} divide the patches on image border into four groups including top, bottom, left and right, and deal with them respectively. In our method, we do not simply consider some certain border in a single direction, but consider the correlations between every two neighboring connective borders. This is considered to be more reasonable since the neighboring connective borders tend to present similar background information. As a result, we get four groups of patches that are located around the image border. Particularly, these groups are denoted as $\{G_q\}_{q=1}^4$, where $G_{1}=B_{t}\bigcup B_{l}, G_{2}=B_{t}\bigcup B_{r}, G_{3}=B_{b}\bigcup B_{l}, G_{4}=B_{b}\bigcup B_{r}$, and $B_{t},B_{b},B_{l},B_{r}$ are the sets of vectorized patches on the top, bottom, left and right borders of the image respectively.

\begin{figure}[t]
\centering
{\includegraphics[width=3.4 in]{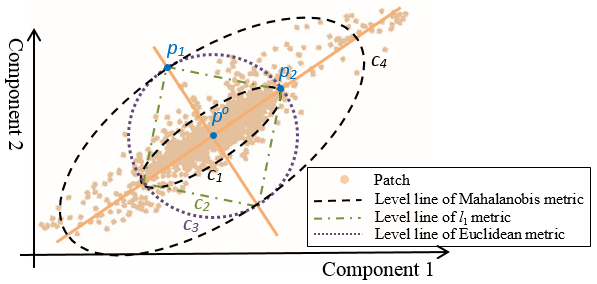}}
\vspace{-5pt}
\caption{The comparison of three metrics to show that \textbf{Mahalanobis distance is more appropriate than Euclidean metric or $l_1$ metric}. The figure shows the distribution of patches of a tested image whose principal components are marked by the orange solid lines. Compared with $p_{2}$, $p_{1}$ is less probable in the background distribution and hence should be considered more distinct, while $p_{2}$ is more probable in the background distribution and hence should not be considered distinct. However, the level lines of $l_1$ metric (green dashdotted line $c_{2}$) and the Euclidean metric (purple dotted line $c_{3}$) reveal the same distance between the two patches to the average patch $p^o$. The level lines of Mahalanobis distance (black dashed line $c_{1}$ and $c_{4}$) show that $p_{1}$ is further away than $p_{2}$ w.r.t $p^o$.}
\label{fig.2}
\end{figure}

To clearly present the proposed idea in the view of metric space, we choose to start with eigendecomposition.
Particularly, with eigendecomposition we have $U_{q}$ and $\Lambda_{q}$ to denote the matrices of eigenvectors and eigenvalues of the covariance matrices $C_q$ for $G_q$, $q=1,\cdots,4$, respectively. With $U_{q}$, for each patch $i$, we have
\begin{eqnarray}\label{1}
t_{q}(i)=U_{q}^{T}f(i), q=1,\cdots,4
\end{eqnarray}
where $t_{q}(i)$ will be the \textbf{element} (new coordinates) of patch $i$ in the $q$-th SBA.

Then, we turn to introduce some appropriate \textbf{metric} for these elements.
Considering the fact that the construction of these elements is carried out on the suspicious background patches, it is speculated that a patch which is similar to the background should be highly probable in the distribution while a patch belongs to the salient region should be highly probable away from the distribution. So, the metric needs to be carefully designed. $l_1$ metric is utilized in \cite{margolin2013makes} to calculate the patch distinctness in its PCA coordinate systems and is proved to be more effective than Euclidean metric. However, it may become invalid in our spaces if one ignores the distribution of the data. In this case, Mahalanobis distance \cite{rahtu2010segmenting}, \cite{li2015adaptive} is used here instead exactly for the elements obtained above, and its effectiveness is explained in detail in Fig. \ref{fig.2}. Therefore, for each patch $i$ we compute a distance $d_{q}(i)$ as its \emph{Mahalanobis distance} via
\begin{eqnarray}\label{3}
d_{q}(i)=\Arrowvert \Lambda_{q}^{-\frac{1}{2}}(t_{q}(i)- p_q^o)\Arrowvert_{2},q=1,\cdots,4
\end{eqnarray}
where $\Lambda_{q}$ is the matrix of eigenvalues and $p_q^o$ is in the transformed space projected from the average patch $f_q^o$ for $G_q$. It is reasonable to understand that $\Lambda_{q}$ used here is to denote the influence of different scales in different components of the eigenvectors from $U_q$  (Fig. \ref{fig.2})  and it is better to describe the distribution of the patches with this metric. In fact, though the distance can be directly computed by $\Arrowvert C_q^{-\frac{1}{2}}(f(i)-f_q^o)\Arrowvert_{2}$, the two ways of definition of Mahalanobis distance are equivalent, and it is more convenient to elaborate the proposed idea by the above context of eigendecomposition.

To suppress the value of $d_q$ from the background as much as possible while retaining the one in the salient regions, we set a threshold H to update $d_{q}$, i.e., if $d_{q}(i)<H$, $d_{q}(i)=0$, where $H$ is set to be the average value of $d_q$ computed by Eq.(\ref{3}). The reason we do this thresholding is originated from the fact that more than 95 percent of images have larger background area than the salient area when investigating a widely used datasets MSRA5000 \cite{liu2011learning}. .

So for each group $G_q$, after repeating the calculations from Eq.(\ref{1}) to Eq.(\ref{3}), we get four \emph{adjusted Mahalanobis distance maps} that labeled as $\{S_{q}^{th}\}_{q=1}^{4}$. Each map is normalized to [0,1]. Next, we create the \emph{single-scale coarse saliency map} $S^{cs}$ by taking a weighted average of the above four maps as
\begin{eqnarray}\label{5}
S^{cs}=\sum_{q=1}^{4}w_{q}\times S_{q}^{th}
\end{eqnarray}
where `$\times$' denotes the element-wise multiplication from here on in, and $w_{q}\in\{0,1\}$. Unlike the
use of entropy in \cite{li2013visual}, we employ it to measure the value of $w_{q}$. Firstly, the entropy values of $\{S_{q}^{th}\}_{q=1}^{4}$ are calculated to be $\{etr_{q}\}_{q=1}^{4}$, respectively. Then the average entropy value $etr_{ave}$ is computed and compared with each $etr_{q}$, that is, if $etr_{q}\le etr_{ave}$, $w_{q}=1$; otherwise, $w_{q}=0$. It means that, for the four maps $\{S_{q}^{th}\}_{q=1}^{4}$, the more chaotic the map is, the less its contribution to $S^{cs}$ will be. In this way, even if some parts of the potential salient object touch part of some border, they will not contribute significantly to the background distribution since only a few  $G_qs$ are influenced, and the SBD in all the $\{G_q\}_{q=1}^4$ may still explore the common features of the true background.

In practice, we calculate $S^{cs}$ on three scales of input, such as 100\%, 50\%, 25\% of the original size of the observed image like the works  \cite{margolin2013makes},\cite{han2014background}, and average them to form the multi-scale  \emph{coarse saliency map} $S^{cm}$.

\begin{figure}[t]
{\includegraphics[width=0.62in]{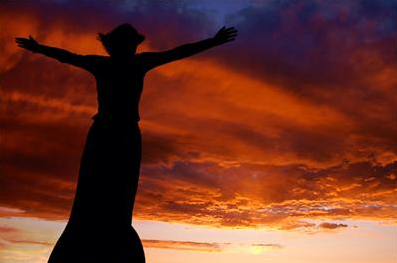}}
{\includegraphics[width=0.62in]{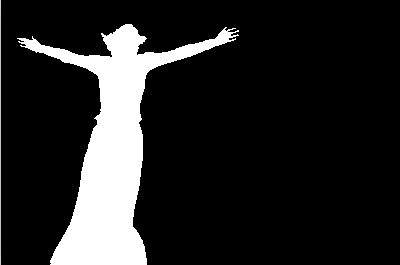}}
{\includegraphics[width=0.62in]{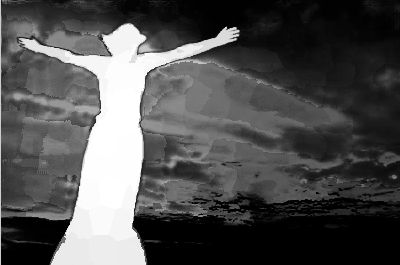}}
{\includegraphics[width=0.62in]{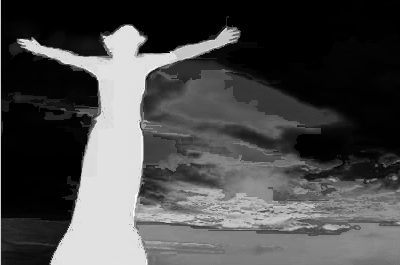}}
{\includegraphics[width=0.62in]{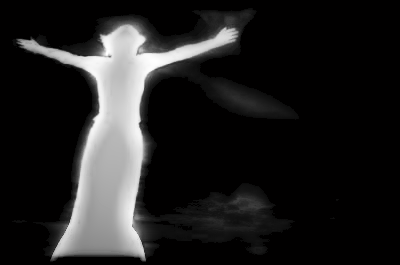}}\\
\par \vspace{-14.pt}
\subfigure[]
{\includegraphics[width=0.62in]{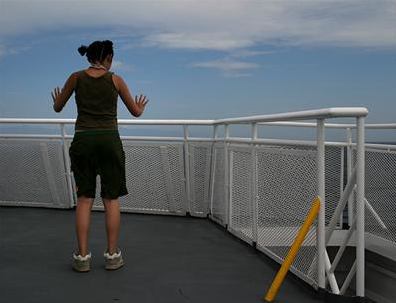}}
\subfigure[]
{\includegraphics[width=0.62in]{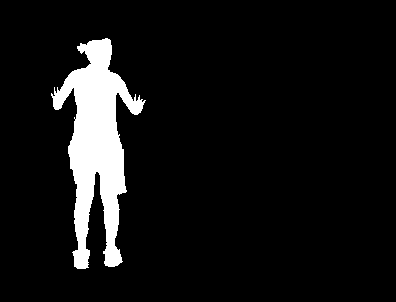}}
\subfigure[]
{\includegraphics[width=0.62in]{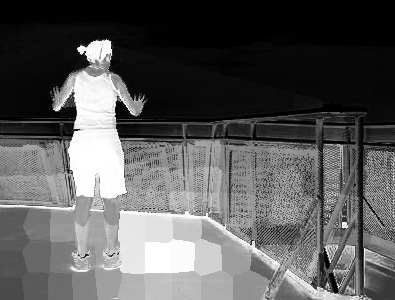}}
\subfigure[]
{\includegraphics[width=0.62in]{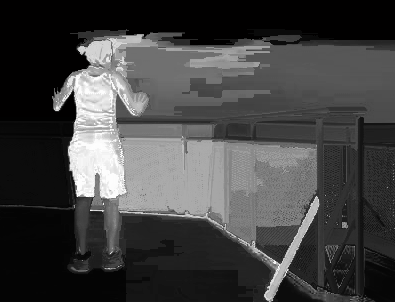}}
\subfigure[]
{\includegraphics[width=0.62in]{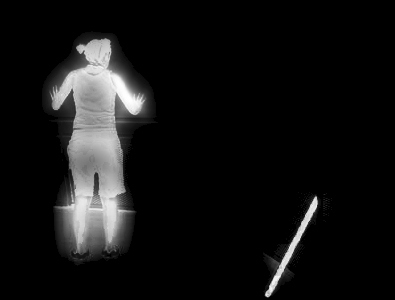}}
\vspace{-7pt}
\caption{Comparisons on different posterior maps within Bayesian scheme. (a) Input images. (b) Ground-truth. (c) \cite{xie2011visual}. (d) \cite{tong2014saliency}. (e) proposed posterior probability maps by Eq.(\ref{9}).}
\label{fig.3}
\end{figure}

\subsection{Bayesian perspective}	
We regard the image saliency as a posterior probability with the Bayes formula to enhance $S^{cm}$ like \cite{xie2011visual} ,\cite{tong2014saliency}. However, unlike these two works where either a convex hull or refined convex hull is used to locate the extracted foreground in determining the \textbf{likelihood}, we extract the foreground region by directly thresholding the \emph{coarse saliency map} $S^{cm}$ with its mean value. Then, the likelihood of each pixel $i$ is defined by \cite{tong2014saliency}
\begin{eqnarray}\label{7}
p(g(i)|R^{1})=\frac{N_{b^{1}(L(i))}}{N_{R^{1}}}\cdot \frac{N_{b^{1}(A(i))}}{N_{R^{1}}}\cdot \frac{N_{b^{1}(B(i))}}{N_{R^{1}}}
\end{eqnarray}
\begin{eqnarray}\label{8}
p(g(i)|R^{0})=\frac{N_{b^{0}(L(i))}}{N_{R^{0}}}\cdot \frac{N_{b^{0}(A(i))}}{N_{R^{0}}}\cdot \frac{N_{b^{0}(B(i))}}{N_{R^{0}}}
\end{eqnarray}
where $g(i)=(L(i),A(i),B(i))$ is the observable vector feature of pixel $i$ from three color channels, i.e.,  $L$, $A$ and $B$ in CIE-Lab color space; $R^1$ (or $R^0$) denotes the extracted foreground (or background); $N_{R^{1}}$ (or $N_{R^{0}}$) denotes the total pixel numbers in $R^1$ (or $R^0$); for channel $L$, $b^{1}(L(i))$ (or $b^{0}(L(i))$) is the bin in pixel-wise color histogram within $R^1$ (or $R^0$) which contains the value $L(i)$, and $N_{b^{1}(L(i))}$ (or $N_{b^{0}(L(i))}$) is the number of pixels in the bin $b^{1}(L(i))$ (or $b^{0}(L(i))$); and it is the same case for channels $A$ and $B$.

Furthermore, we treat the \emph{coarse saliency map} $S^{cm}$ from Section A as the \textbf{prior probability} map, and calculate the \textbf{posterior probability} map for pixel $i$ by
\begin{eqnarray}\label{9}
S_{p}(i)=\frac{p(g(i)|R^{1})\cdot S^{cm}(i)}{p(g(i)|R^{1})\cdot S^{cm}(i)+p(g(i)|R^{0})\cdot (1-S^{cm}(i))}
\end{eqnarray}

As shown in Fig. \ref{1} (c), $S^{cm}$ are strongly enhanced. Besides, we compare the $S_{p}$ generated by Eq.(\ref{9}) with other posterior probability maps in \cite{xie2011visual},\cite{tong2014saliency}, and present the results in Fig. \ref{fig.3} to show that the proposed way performs better to both suppress the background noises and highlight the salient regions.

\begin{figure}[t]
{\includegraphics[width=0.64in]{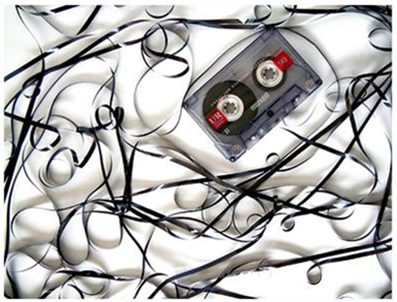}}
{\includegraphics[width=0.64in]{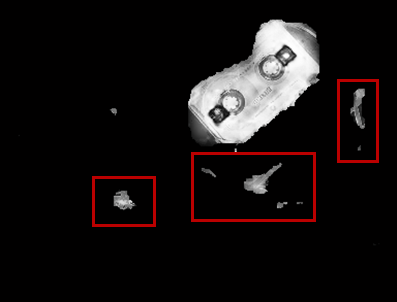}}
{\includegraphics[width=0.64in]{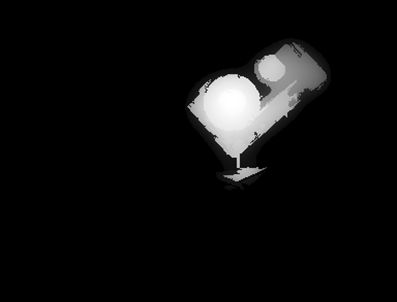}}
{\includegraphics[width=0.64in]{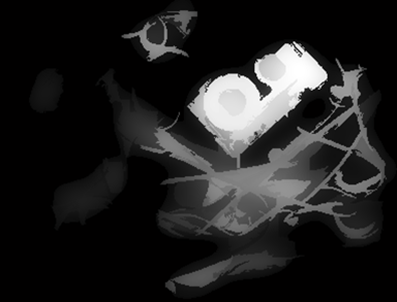}}
{\includegraphics[width=0.64in]{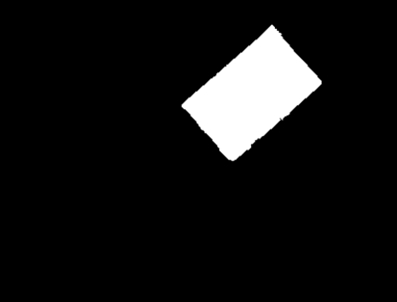}}\\
\par \vspace{-14.pt}
\subfigure[]
{\includegraphics[width=0.64in]{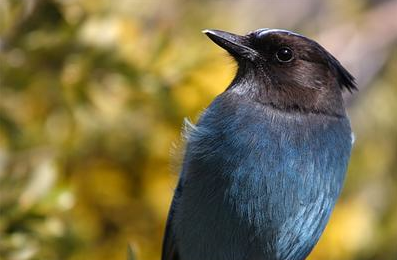}}
\subfigure[]
{\includegraphics[width=0.64in]{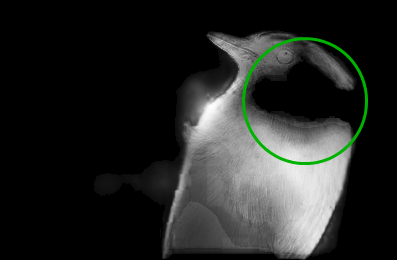}}
\subfigure[]
{\includegraphics[width=0.64in]{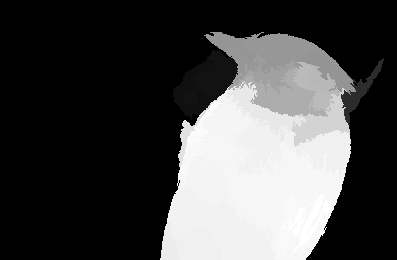}}
\subfigure[]
{\includegraphics[width=0.64in]{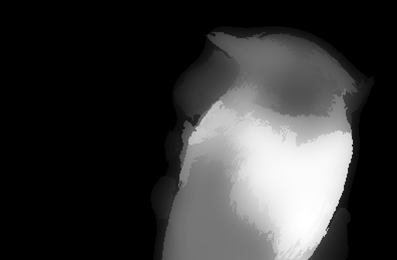}}
\subfigure[]
{\includegraphics[width=0.64in]{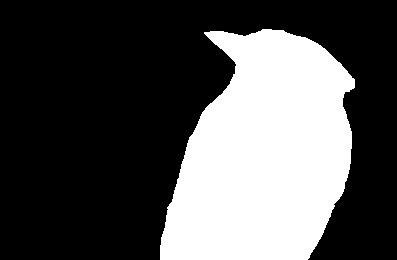}}
\vspace{-7pt}
\caption{Two possible misjudged situations after Bayes enhancement, which are shown in the background region (red box) and inside the salient object (green circle). (a) Input images. (b) The posterior probability maps by Eq.(7). (c) Saliency maps based on the proposed refinement with geodesic distance. (d) Saliency maps based on the up-sampling method in \cite{perazzi2012saliency}. (e) Ground-truth.}
\label{fig.4}
\end{figure}
\vspace{-3pt}

\begin{figure*}[t]
\centering
\subfigure[]
{\includegraphics[width=1.7in]{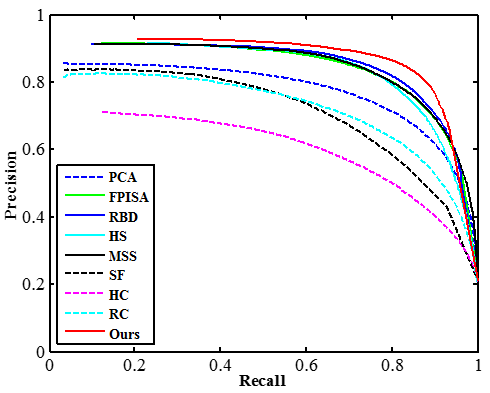}}
\subfigure[]
{\includegraphics[width=1.7in]{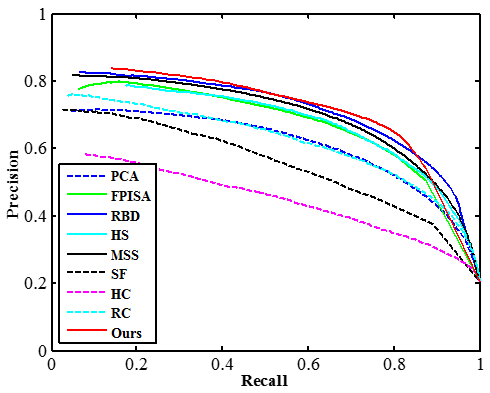}}
\subfigure[]
{\includegraphics[width=1.7in]{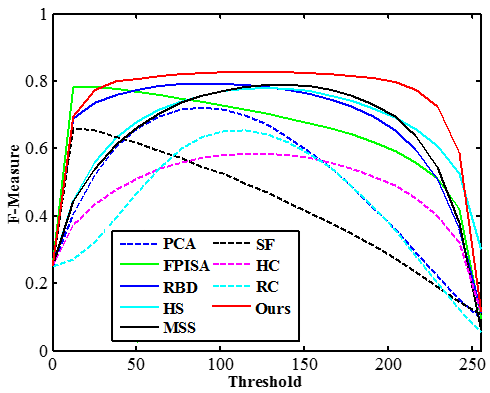}}
\subfigure[]
{\includegraphics[width=1.7in]{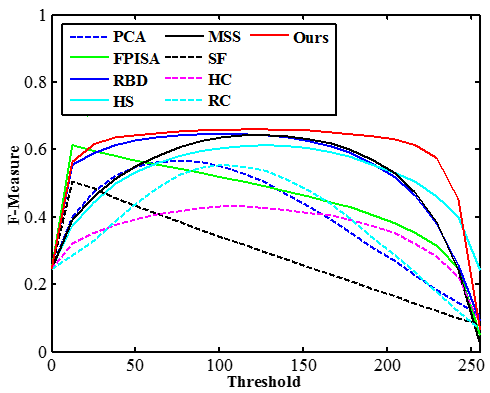}}
\vspace{-7.pt}
\caption{PR curves and F-measure curves for different methods. (a) PR curves on MSRA5000 dataset. (b) PR curves on PASCAL1500 dataset. (c) F-measure curves on MSRA5000 dataset. (d) F-measure curves on PASCAL1500 dataset.}
\label{fig.5}
\end{figure*}

\begin{table*}[t]
 \centering\
 \begin{threeparttable}
 \caption{\label{tab:results}Quantitative Results For Different Methods. }
  \begin{tabular}{|*{11}{c|}}
  \hline
  \multicolumn{2}{|l|}{} &Ours &RBD &FPISA &MSS &HS &PCA &SF &HC &RC\\
  \hline
    \multirow{5}*{MSRA5000}
   &Precision(AT) &\textbf{.8105} &.7824 &.7979 &.6256 &.6200 &.5569 &.7494 &.4706 &.4033 \\
   &Recall(AT) &\textbf{.7269} &.6470 &.4991 &.6959 &.6940 &.5117 &.2976 &.5644 &.5628 \\
   &F-measure(AT) &\textbf{.7696} &.7222 &.6688 &.6260 &.6156 &.5262 &.5079 &.4719 &.4174 \\
   \cline{2-11}
   &AUC &\textbf{.9192} &.9127 &.8923 &.9189 &.8988 &.9022 &.8616 &.7701 &.8633 \\
   \cline{2-11}
   &MAE &\textbf{.0927} &1106 &.1307 &.1489 &.1620 &.1888 &.1660 &.2391 &.2638 \\
  \hline
   \multirow{5}*{PASCAL1500}
   &Precision(AT) &\textbf{.6682} &.6343 &.6650 &.5223 &.4923 &.4840 &.6058 &.3653 &.3767 \\
   &Recall(AT) &\textbf{.6488} &.5724 &.3306 &.6006 &.6243 &.4323 &.1901 &.4879 &.5226 \\
   &F-measure(AT) &\textbf{.6231} &.5749 &.4754 &.5092 &.4851 &.4270 &.3491 &.3613 &.3768 \\
   \cline{2-11}
   &AUC &.8676 &\textbf{.8728} &.8144 &.8723 &.8433 &.8549 &.8088 &.7119 &.8436 \\
   \cline{2-11}
   &MAE &\textbf{.1420} &.1539 &.2223 &.1861 &.1739 &.2092 &.1920 &.2927 &.2712 \\
  \hline
  \end{tabular}
 \end{threeparttable}
\end{table*}


\subsection{Refinement with Geodesic Distance}
Additionally, there still exist two misjudged situations may occur after the above steps. For example in  Fig. \ref{fig.4} (b), either some of the pixels from background or from the salient object are wrongly judged. For refinement, like \cite{perazzi2012saliency}, we adopt an up-sampling method where we suppose that one pixel's saliency value is determined by a weighted linear combination of the saliency of its surrounding parts. However, unlike \cite{perazzi2012saliency} which suggests that the weights be influenced by the color and the position information of the surround pixels, we consider that these weights should be sensitive to a \emph{geodesic distance}. The \textbf{input image} is firstly segmented into a number of superpixels based on Linear Spectral Clustering (LSC) method \cite{Li_2015_CVPR},\cite{perazzi2012saliency} and the posterior probability of each superpixel is calculated by averaging the posterior probability values $S_{p}$ of all its pixels inside. Then for $j$th superpixel, if its posterior probability is  labeled as $\bar{S}(j)$, thus the saliency value of the $q$-th superpixel is measured by
\begin{eqnarray}\label{10}
S(q)=\sum_{j=1}^{N}w_{qj} \cdot \bar{S}(j)
\end{eqnarray}
where $N$ is the total number of superpixels, and $w_{qj}$ will be a weight based on the \emph{geodesic distance}\cite{zhu2014saliency} (to be quoted) between superpixel $q$ and $j$.
Similar to \cite{zhu2014saliency}, we firstly construct an undirected weighted graph by connecting all adjacent superpixels $(a_{k},a_{k+1})$ and assigning their weight $d_{col}(a_{k},a_{k+1})$ as the Euclidean distance between their average colors in the CIE-Lab color space. Then the \emph{geodesic distance} between any two superpixels $d_{geo}(q,j)$ is defined as the
accumulated edge weights along their shortest path on the graph via
\begin{eqnarray}\label{11}
d_{geo}(q,j)=\underset{a_{1}=q,a_{2},\cdots ,a_{n}=j}{min}\sum_{k=1}^{n-1}d_{col}(a_{k},a_{k+1}),
\end{eqnarray}

Next we define the weight $w_{qj}$ between two superpixels $(q,j)$ as $w_{qj}=exp(-\frac{d_{geo}^{2}(q,j)}{2\sigma_{col}^{2}})$. We can see that, when $q$ and $j$ are in a flat region, $d_{geo}(q,j)=0$ and $w_{qj}=1$, ensuring that saliency of $j$ has high contribution to the saliency of $q$; when $q$ and $j$ are in different regions, there exists at least one strong edge ( $d_{col}(*, *)\ge 3\sigma_{col}$ ) on their shortest path and $w_{qj}\approx 0$, ensuring that the saliency of $j$ should not contribute to the saliency of $q$.  $\sigma_{col}$ is the deviation for all $d_{col}(*, *)$ and computed in practice. Fig. \ref{fig.4} (c) and (d) show the comparisons based on our refinement methods and the up-sampling proposed in \cite{perazzi2012saliency} and our method performs better. 

\section{Experiment}
We evaluate and compare with other methods the performance of the proposed algorithm on two widely used datasets: MSRA5000 \cite{liu2011learning} and PASCAL1500 \cite{wang2015pisa}. The MSRA5000 contains 5000 images with pixel-level ground-truths, while the PASCAL1500 which contains 1500 images with pixel-level ground-truths is more challenging for testing because they have more complicated patterns in both foreground and background. Then, we compare our results with eight state-of-the art methods, including HC \cite{cheng2015global}, RC \cite{cheng2015global}, SF \cite{perazzi2012saliency}, PCA \cite{margolin2013makes}, FPISA \cite{wang2015pisa}, HS \cite{yan2013hierarchical}, MSS \cite{tong2014saliency}, RBD \cite{zhu2014saliency}, among which are discussed and evaluated in the benchmark paper \cite{borji2015salient}.

\begin{figure}[t]
{\includegraphics[width=0.54in]{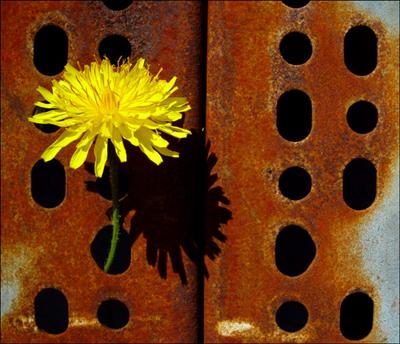}}
{\includegraphics[width=0.54in]{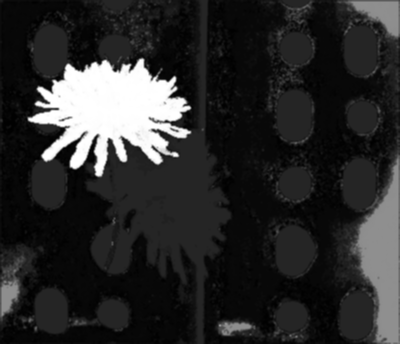}}
{\includegraphics[width=0.54in]{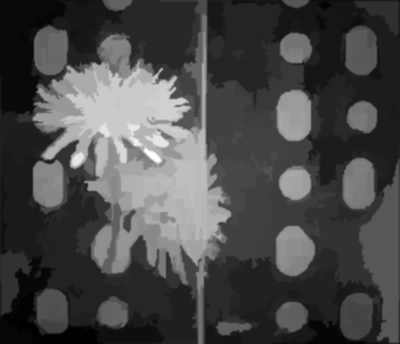}}
{\includegraphics[width=0.54in]{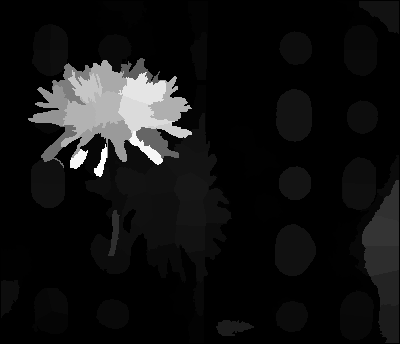}}
{\includegraphics[width=0.54in]{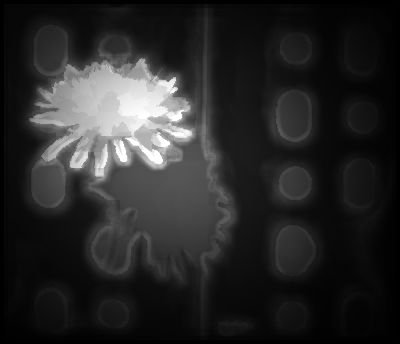}}
{\includegraphics[width=0.54in]{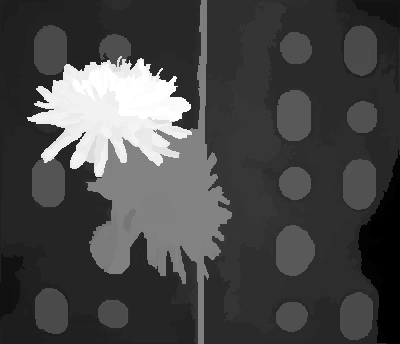}}
\par \vspace{3.pt}
{\includegraphics[width=0.54in]{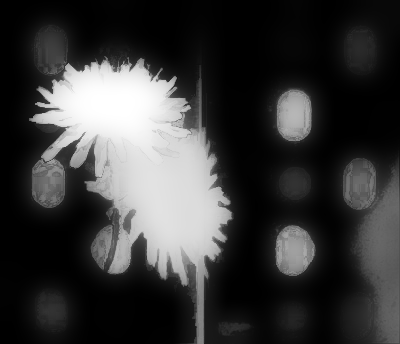}}
{\includegraphics[width=0.54in]{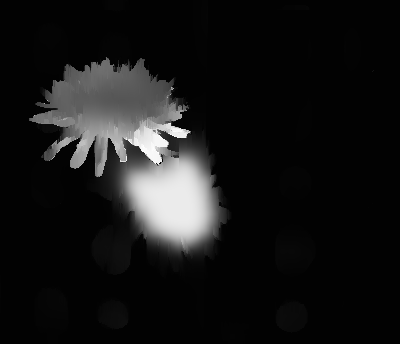}}
{\includegraphics[width=0.54in]{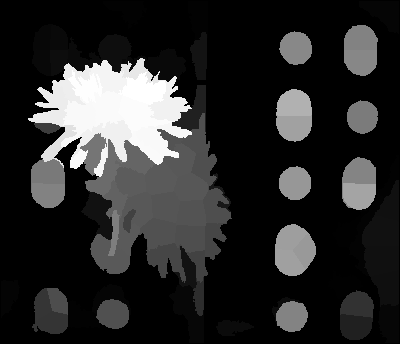}}
{\includegraphics[width=0.54in]{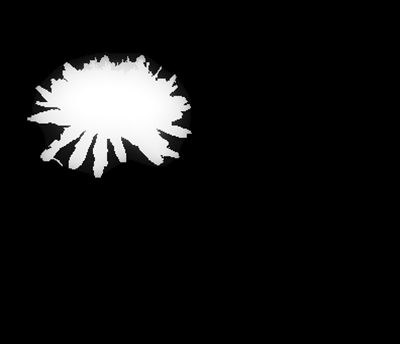}}
{\includegraphics[width=0.54in]{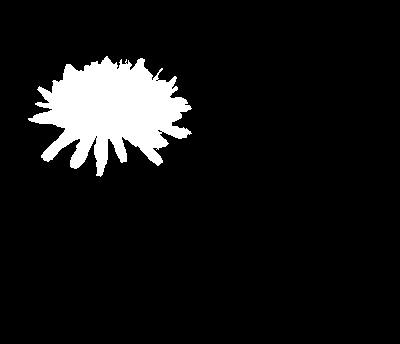}}
\vspace{-2pt}
\caption{Visual comparison of saliency maps of different methods, where from left to right, top to down are : input, HC \cite{cheng2015global}, RC \cite{cheng2015global}, SF \cite{perazzi2012saliency}, PCA \cite{margolin2013makes}, HS \cite{yan2013hierarchical}, MSS \cite{tong2014saliency}, FPISA \cite{wang2015pisa}, RBD \cite{zhu2014saliency}, ours and ground-truth.}
\label{fig.6}
\end{figure}

In the study, we use 4 standard criteria for quantitative evaluation, i.e., precision-recall (PR) curve \cite{tong2014saliency},  F-measure \cite{yan2013hierarchical}, Mean Absolute Error (MAE) \cite{perazzi2012saliency},\cite{zhu2014saliency} and AUC score \cite{margolin2013makes},\cite{tong2014saliency}.  For the PR curve, it is given in Fig. \ref{fig.5}(a) and (b) by comparing the results with the ground-truth through varying thresholds within the range [0, 255]. And F-measure curve in Fig. \ref{fig.5}(c) and (d) is drawn with different $F_{\beta}$ values calculated from the adaptive thresholded saliency map, where $F_{\beta}=\frac{(1+\beta^{2}) precision \cdot recall}{\beta^{2}\cdot precision+recall}$, and $\beta^{2}=0.3$. The figures show that our approach achieves the better for these two measures. Specially, similar to \cite{perazzi2012saliency}, \cite{wang2015pisa}, in TABLE \ref{tab:results} we show the precision, recall and F-measure values for adaptive threshold (AT), which is defined as twice the mean saliency of the image. It still shows that the proposed method achieves the best on these data.  In addition, MAE is used to evaluate the averaged degree of the dissimilarity and comparability between the saliency image and the ground-truth at every pixel, and AUC score is calculated based on the true positive and false positive rates. It is shown in TABLE \ref{tab:results} that the proposed method presents the smallest MAE to denote our saliency maps are more close to the ground-truth at pixel level, and also offers satisfactory AUC scores compared with the others. Finally, Fig. \ref{fig.6} shows series of saliency maps of different methods. And it shows that our result has a great improvement over previous methods. 
W.r.t. the computational efficiency, the above average runtime for an input image is 1.67s in Matlab platform on a PC with Intel i5-4460 CPU and 8GB RAM.

\vspace{-2pt}
\section{Conclusion}
In this letter, an effective bottom-up saliency detection method with the so-called spaces of background-based distribution (SBD) is presented. In order to describe the background distribution, some metric spaces of SBD are constructed, i.e., the elements are generated with eigendecomposition, plus the metric is designed carefully where the Mahalanobis distance is quoted to measure the saliency value for each patch. Further by enhancement in Bayesian perspective and refinement with geodesic distance, the whole salient detection is done. Finally, the evaluational experiments exhibit the effectiveness of the proposed algorithm compared with the state-of-art methods.

\ifCLASSOPTIONcaptionsoff
\newpage
\fi


\end{document}